\documentclass[runningheads]{llncs}
\usepackage[T1]{fontenc}
\usepackage{graphicx}
\usepackage{amsmath}
\usepackage{amsfonts}
\usepackage{subcaption}
\usepackage{booktabs}
\usepackage[x11names,dvipsnames,table]{xcolor} 
\usepackage{colortbl}
\usepackage{hyperref}
\usepackage{pdfpages}

\begin{document}
%
\title{Beyond the Vehicle: Cooperative Localization by Fusing Point Clouds for GPS-Challenged Urban Scenarios}
\titlerunning{Beyond the Vehicle}
%
\author{Kuo-Yi Chao\inst{1}\orcidID{0000-0002-1122-4072}, Ralph Rasshofer\inst{2} \and \\
Alois Christian Knoll\inst{1}\orcidID{0000-0003-4840-076X}}
\authorrunning{K. Chao et al., 2025}
%
\institute{$^1$Technical University of Munich,\\Chair of  Robotics, Artificial Intelligence (AI) and Embedded Systems, Boltzmannstraße 3, 85748 Garching bei München, Germany\\
\email{kuoyi.chao@tum.de, knoll@tum.de}\ \\
$^2$BMW Group, Munich, Germany\\
\email{ralph.rasshofer@bmw.de}\
}

\maketitle              
\begin{abstract}
Accurate vehicle localization is a critical challenge in urban environments where GPS signals are often unreliable. This paper presents a cooperative multi-sensor and multi-modal localization approach to address this issue by fusing data from vehicle-to-vehicle (V2V) and vehicle-to-infrastructure (V2I) systems. Our approach integrates cooperative data with a point cloud registration-based simultaneous localization and mapping (SLAM) algorithm. The system processes point clouds generated from diverse sensor modalities, including vehicle-mounted LiDAR and stereo cameras, as well as sensors deployed at intersections. By leveraging shared data from infrastructure, our method significantly improves localization accuracy and robustness in complex, GPS-noisy urban scenarios.

\end{abstract}
\section{Introduction}
Accurate and robust localization is a cornerstone of safe autonomous driving, enabling vehicles to precisely determine their position in complex environments \cite{herau2025poseoptimizationautonomousdriving} \cite{localization}. Conventional methods relying solely on Global Navigation Satellite Systems (GNSS) often suffer from significant errors in urban canyons or adverse weather conditions \cite{gps-error}. To overcome these limitations, cooperative multi-modal localization has emerged as a promising solution. This approach leverages Vehicle-to-Everything (V2X) communication to fuse data from a vehicle's own sensors, such as LiDAR and cameras, with data from surrounding vehicles and roadside infrastructure.

The use of 3D point clouds, generated from LiDAR or stereo camera disparity maps, provides rich geometric details of the environment. Integrating these point clouds into Simultaneous Localization and Mapping (SLAM) \cite{rs13224497} frameworks has been shown to enhance positioning accuracy. However, much of the existing work focuses primarily on Vehicle-to-Vehicle (V2V) data exchange. The systematic integration of infrastructure-based point clouds to augment vehicle localization remains a key challenge.

In this paper, we propose a novel pipeline that merges point clouds from multiple vehicles and intersection-based sensors. By fusing these distributed datasets, our approach achieves highly accurate and robust cooperative localization, even in scenarios where GNSS \cite{11028497} is unreliable.

\section{Background and Methodology}
Cooperative localization leverages V2X communication to share sensory and positional data among vehicles and infrastructure. Multi-modal fusion integrates data at data, feature, or decision levels, combining raw sensor data, extracted features, or high-level decisions.

\begin{figure}
\centering
\includegraphics[width=\linewidth]{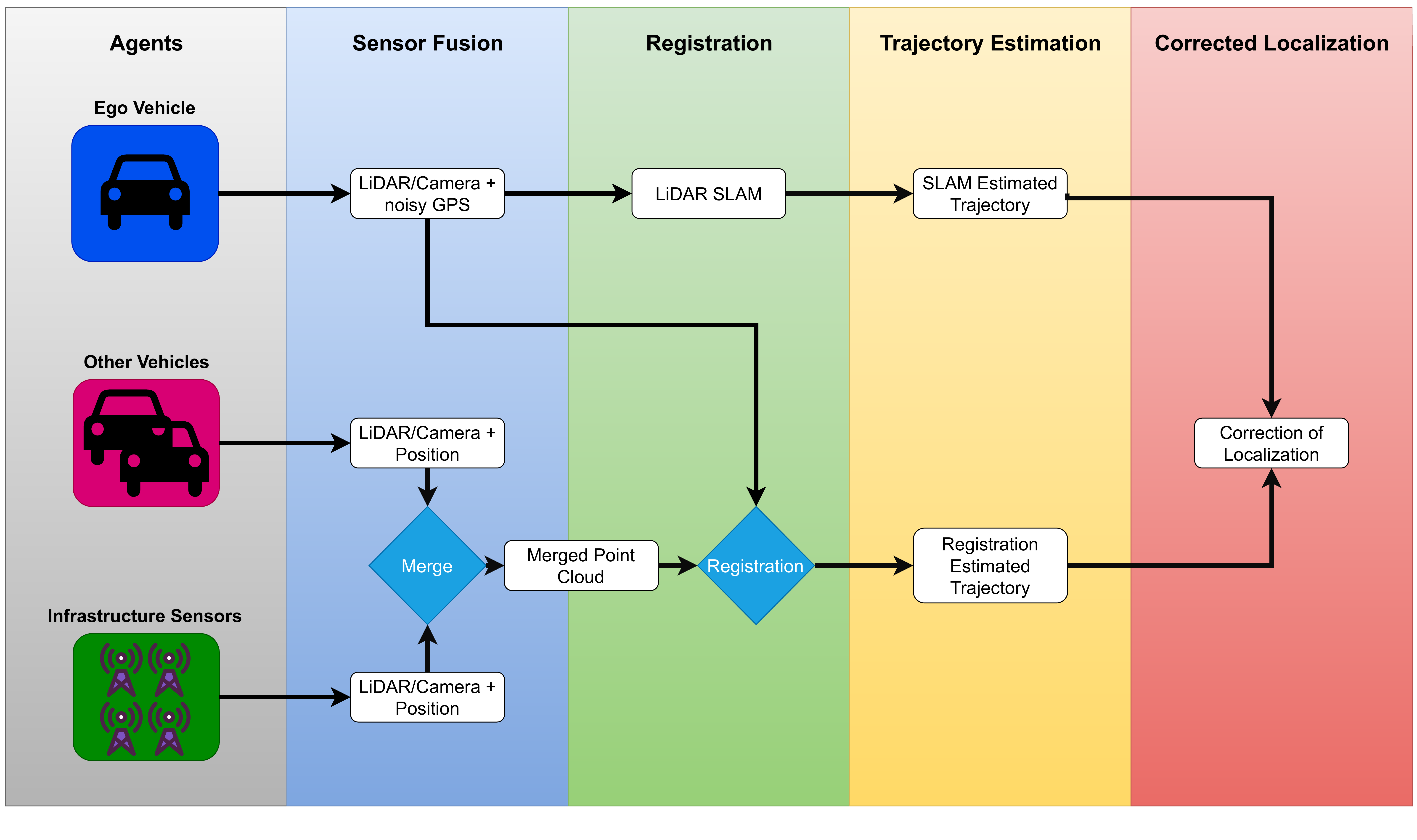}
\caption{Pipeline for sensor fusion to correct noisy GPS data. The pipeline contains 4 steps to correct its localization by using external sensor information from other agents: sensor fusion, registration, trajectory estimation, and finally correction.}
\label{fig1:main-pipeline}
\end{figure}

In our approach, the ego vehicle initially relies on its onboard sensors, such as LiDAR and mono/stereo cameras \cite{sensor}, to perform Simultaneous Localization and Mapping (SLAM) \cite{rs15041156}, which estimates the vehicle's trajectory and builds a local map of the environment. Over time, however, SLAM accumulates drift \cite{KEITAANNIEMI2023104700}, particularly exacerbated by inaccuracies in GPS data from multipath interference, mentioned in \autoref{source_gps_error}, or signal obstructions in urban settings \cite{gentner2016slam}. This problem occurs in real-world scenarios; thus, an online SLAM is needed, as described in \autoref{ch:online_slam}. For point cloud fusion, we incorporate environmental information from intersection-based infrastructure (e.g., Intelligent Traffic Systems) \cite{10375912} and sensors on nearby vehicles, aggregating their point clouds into a comprehensive reference map. In this map, we use Fast Point Feature Histograms as descriptors, mentioned in \autoref{ch:fpfh}. This fused data is then shared with the ego vehicle, enabling Iterative Closest Point (ICP) \cite{924423}, mentioned in \autoref{ch:icp}, registration augmented with Random Sample Consensus (RANSAC) \cite{FISCHLER1987726}, \autoref{ch:registration}, for robust outlier rejection and alignment. By registering the ego vehicle's local point clouds against this global reference, the framework corrects localization errors, achieving enhanced accuracy and reduced drift in real-time navigation.
\subsection{Sources of GPS Error}\label{source_gps_error}
GPS positioning in semi-urban environments faces amplified errors over open-sky conditions \cite{GPS}. Key factors include:
\begin{itemize}
\item \textbf{Multipath Interference} \cite{GPS} \cite{10578540} \cite{ublox2023multipath} \cite{gentner2016slam}: Signals reflect off buildings or vehicles, causing pseudo-range errors via indirect paths:
\begin{equation}
    \Delta d_{\text{mult}} = \sum \alpha_k (d_k - d_0 + \phi_k),
\end{equation}

where $0 < \alpha_k \leq 1$ is attenuation and $\phi_k$ is phase shift.
\item \textbf{Atmospheric Delays} \cite{rs12213569} \cite{s8128479}: Ionospheric and tropospheric delays, with ionospheric part:
\begin{equation}
    \Delta d_{\text{ion}} = 40.3 \cdot \frac{\text{TEC}}{f^2}
\end{equation}

meters ($f$ in GHz, Total Electron Content (TEC) in TEC Unit (TECU).
\item \textbf{Urban Canyon Effects} \cite{urban}: Structures block satellites, worsening geometry and Position Dilution of Precision (PDOP), amplifying noise.
\end{itemize}

\subsection{Online SLAM} \label{ch:online_slam}
Online Simultaneous Localization and Mapping (SLAM) \cite{1507422} involves real-time algorithms for autonomous agents (e.g., vehicles) to incrementally map unknown environments while estimating their position, processing live sensor data like LiDAR or cameras without batch processing. Unlike offline SLAM, which optimizes full datasets retrospectively, online SLAM is vital for dynamic scenarios like semi-urban navigation needing instant GPS corrections. For example, PIN-SLAM \cite{pan2024tro} uses point-based implicit neural representations for odometry, loop closure, and adaptive mapping to boost accuracy and efficiency \cite{herau2025poseoptimizationautonomousdriving}. Mathematically, it relies on Bayesian inference to compute the posterior $   p\bigl(\mathbf{x}_{1:t},\,\mathbf{m}\;\bigm|\;\mathbf{z}_{1:t},\,\mathbf{u}_{1:t}\bigr)
 \bigl(\text{ with trajectory }\mathbf{x}_{1:t},\;\text{map }\mathbf{m},\;
         \text{observations }\mathbf{z}_{1:t},\;\text{controls }\mathbf{u}_{1:t}\bigr)$, approximated via methods like Extended Kalman Filter (EKF) \cite{EKF} for linearizing models and updating estimates, or graph-based optimization minimizing errors in pose graphs for loop closures \cite{stathoulopoulos2025balancingaccuracyefficiencylargescale}.

\subsection{Fast Point Feature Histograms}\label{ch:fpfh}
In this work, we computed Fast Point Feature Histograms (FPFH) as a robust local descriptor \cite{OguzDGKP24} for 3D point clouds, enhancing correspondence estimation in registration tasks \cite{5152473}, following a downsampling step to reduce computational complexity while preserving key geometric features. This downsampling process simplifies the FPFH calculation by focusing on a subset of points, where FPFH captures the relationships between a query point and its neighbors, generating a 33-dimensional histogram that encodes surface normals, curvatures, and angular deviations.

\subsection{Point Cloud Registration} \label{ch:registration}
Point cloud registration is the process of finding the optimal rigid transformation ($\mathbf{T} \in SE(3)$) that aligns two or more 3D point clouds into a common coordinate frame \cite{huang2021comprehensivesurveypointcloud} \cite{arnold2021fastrobustregistrationpartially}. In this work, we use registration to enhance vehicle localization by aligning the ego vehicle's LiDAR point cloud with a merged reference map generated from nearby sensors with known locations. The registration is done by the iterative closest point (ICP) algorithm \cite{Bai_2023} \cite{rs14194874}.

Given a source point cloud $\mathcal{P}_s = \{\mathbf{p}_i^s \in \mathbb{R}^3\}$ and a target point cloud $\mathcal{P}_t = \{\mathbf{p}_j^t \in \mathbb{R}^3\}$, the goal is to find the rotation $\mathbf{R} \in SO(3)$ and translation $\mathbf{t} \in \mathbb{R}^3$ that minimize the distance between corresponding points \cite{Li_2024} $
     (\mathbf{R_{k+1}}, \mathbf{t_{k+1}})=\arg\min_{\mathbf{R}, \mathbf{t}} \sum_{i=1}^N \left\| \mathbf{R} \mathbf{p}_i^s + \mathbf{t} - \mathbf{p}_i^t \right\|^2,$
where the correspondences ${\mathbf{p}_i^s, \mathbf{p}_i^t}$ are assumed known or estimated. In practice, this optimization problem is solved in two stages: coarse (global) registration to find an approximate alignment, followed by fine (local) registration to refine.

\subsubsection{RANSAC} \label{ch:icp}
The classic ICP algorithm risks local minima from poor initialization and outlier sensitivity to noise/occlusions. Random sample consensus (RANSAC) \cite{FISCHLER1987726} addresses this via robust sampling of minimal subsets (e.g., 3 - 4 points), model fitting, inlier identification within threshold $\epsilon$, and best-model selection. The ICP-RANSAC \cite{f15060893} hybrid boosts accuracy in noisy cases like multi-agent fusion for semi-urban GPS trajectory correction.

\section{Experiments and Results} \label{experiments}
\begin{figure}[ht]
    \centering
    \begin{subfigure}
    {0.48\textwidth}
    \begin{center}
        
    \includegraphics[width=\linewidth]{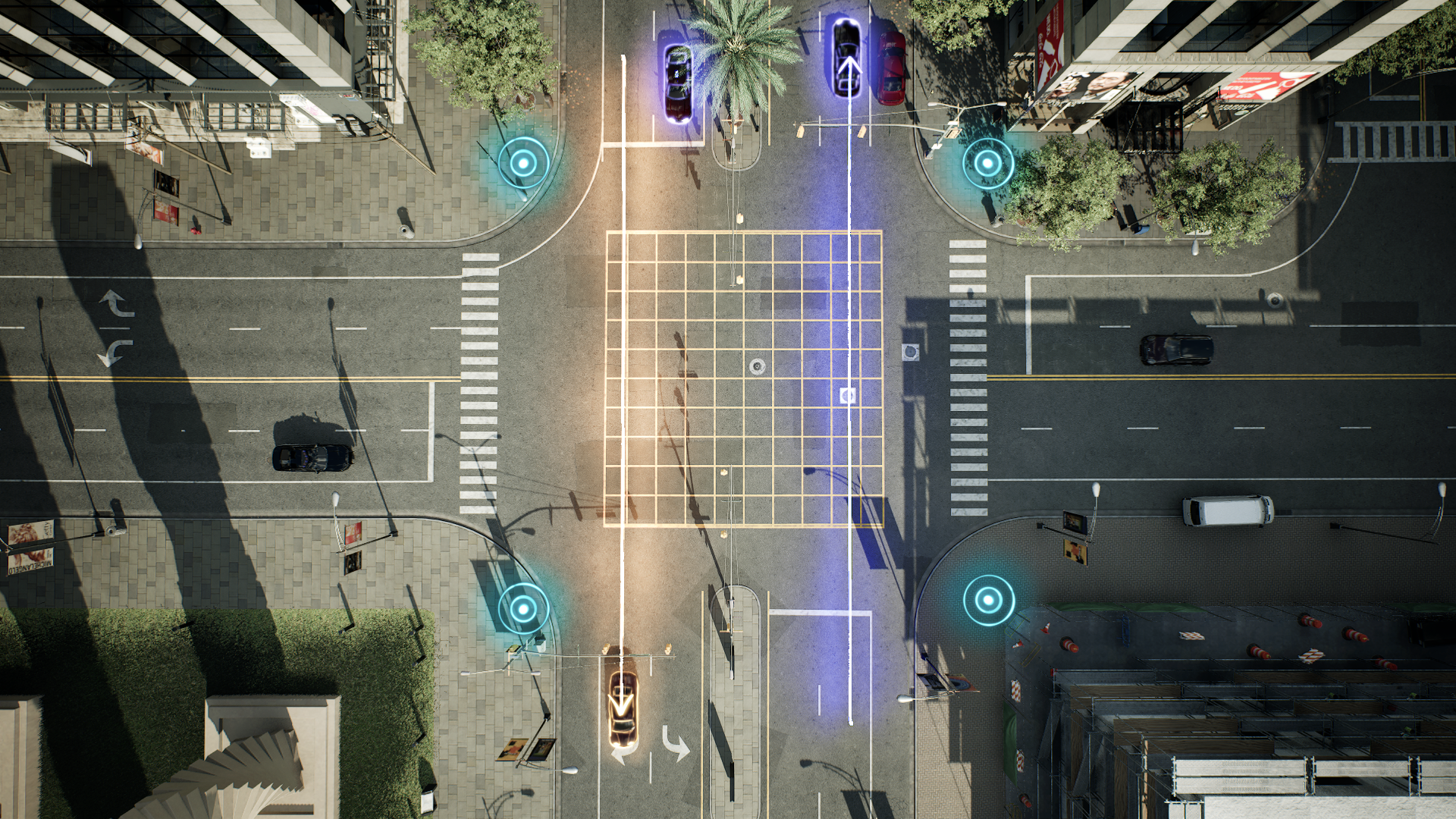}
        \caption{Sim 0}
        \end{center}
    \end{subfigure}
    \hfill
    \begin{subfigure}{0.48\textwidth}
        \includegraphics[width=\linewidth]{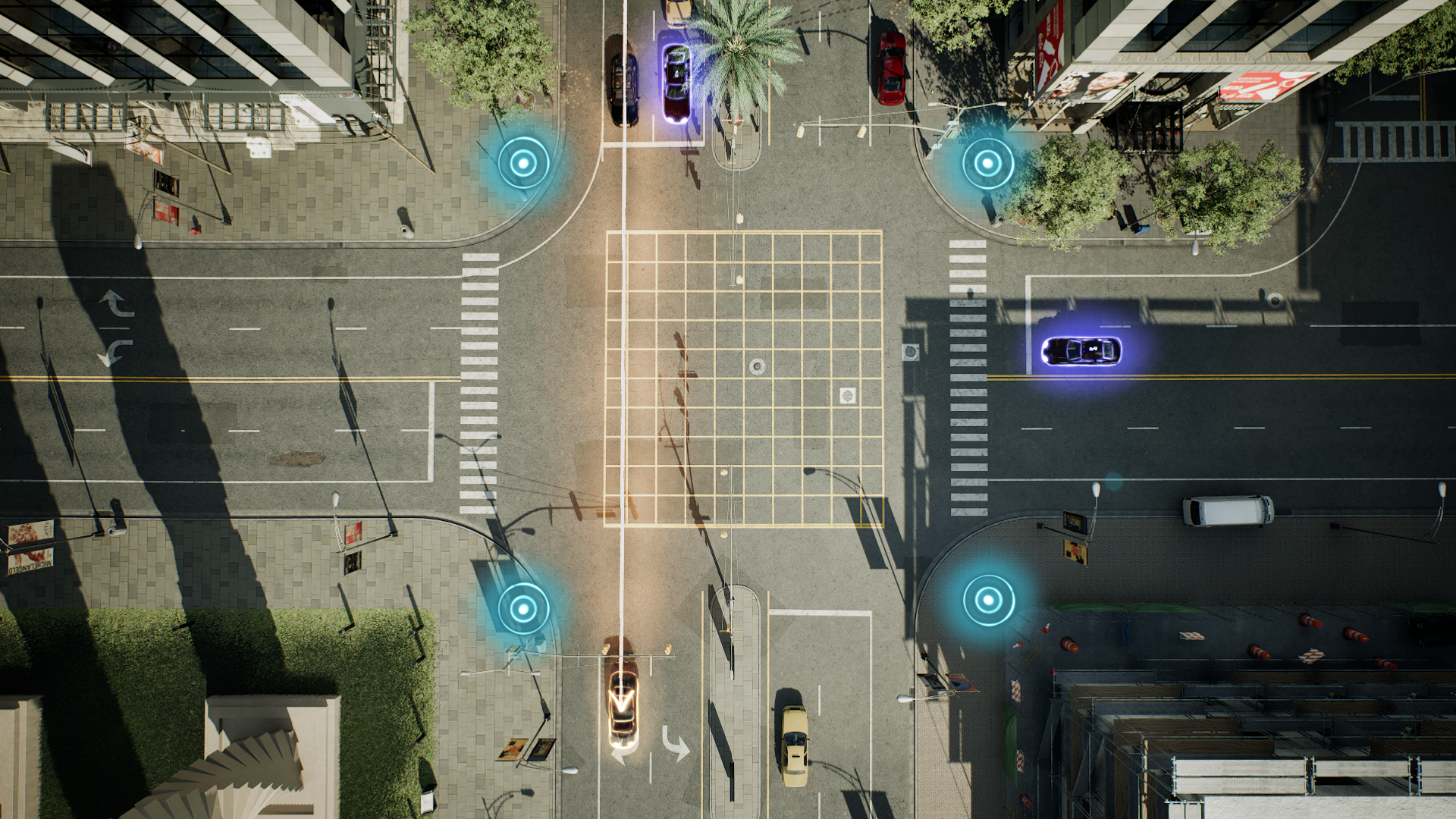}
        \caption{Sim 1}
    \end{subfigure}

    \vspace{1em}

    \begin{subfigure}{0.48\textwidth}
        \includegraphics[width=\linewidth]{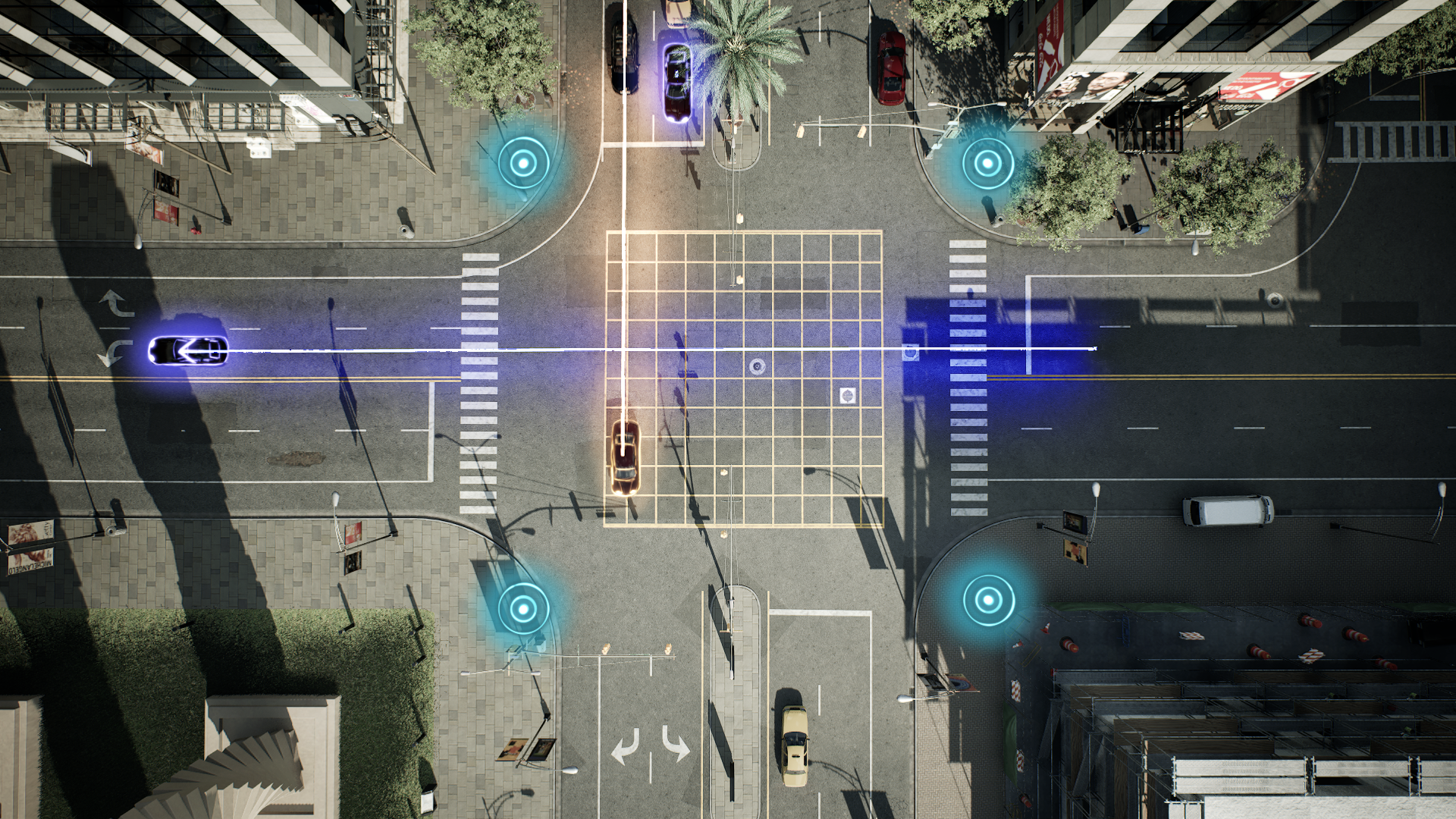}
        \caption{Sim 2}
    \end{subfigure}
    \hfill
    \begin{subfigure}{0.48\textwidth}
        \includegraphics[width=\linewidth]{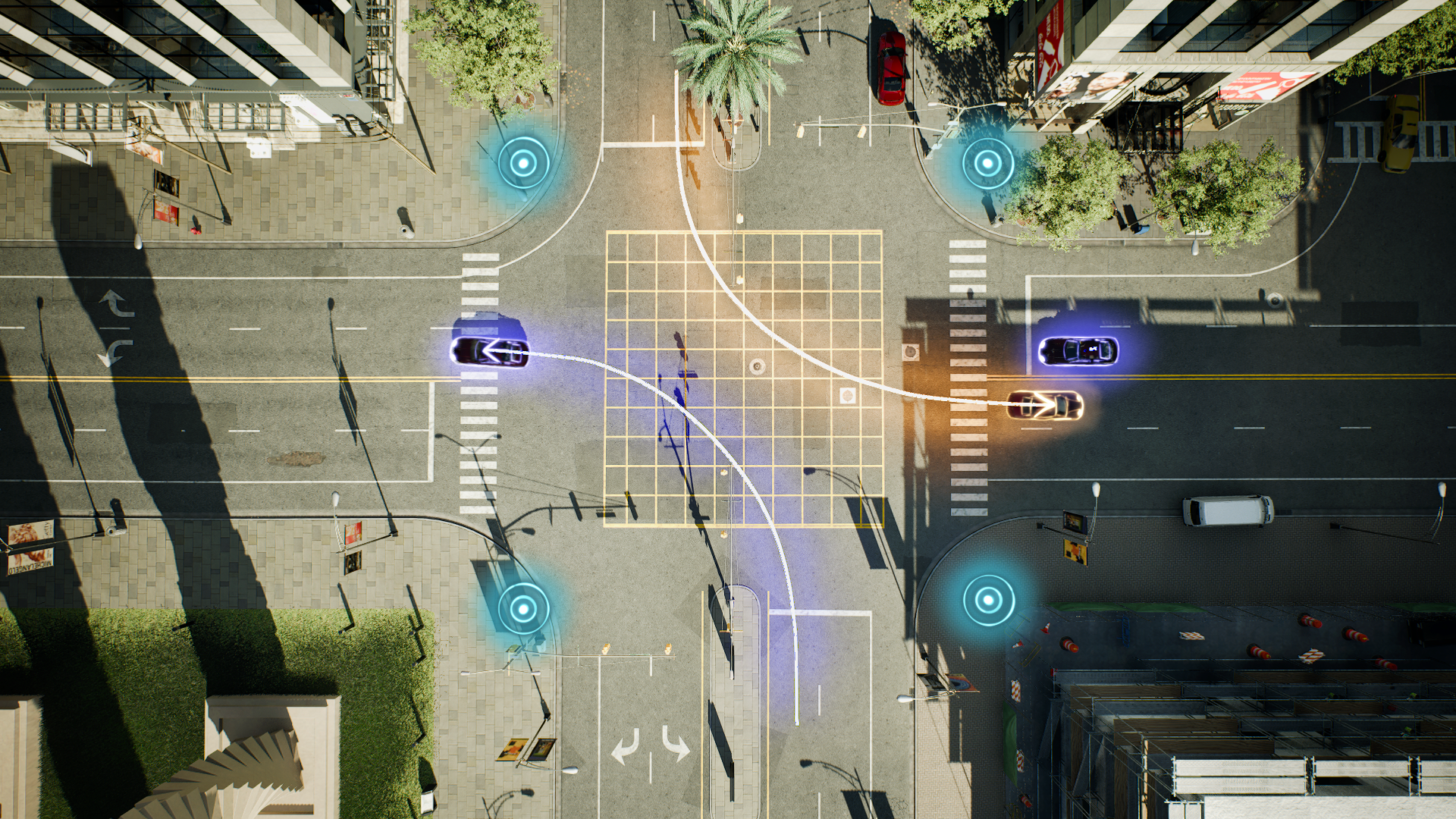}
        \caption{Sim 3}
    \end{subfigure}

    \caption{Simulation screenshots \cite{CARLA} from each scenario at a single intersection, displaying the ego vehicle (orange path), two agent vehicles (blue paths/dots), and four infrastructure sensors (aqua dots) at corners for traffic monitoring. Sim~0: ego crosses parallel opposing paths with one agent; Sim 1: agents stationary; Sim 2: ego crosses perpendicular paths with one agent; Sim 3: ego and one agent execute left turn.}
    \label{fig:sim-grid}
\end{figure}

We set up a CARLA simulation \cite{CARLA} environment, as illustrated in Fig. \ref{fig:sim-grid}, equipping the ego vehicle with one LiDAR and two cameras. The intersection features multiple LiDAR sensors, while each agent vehicle is fitted with one LiDAR. Various scenarios were implemented to evaluate the setup.

Tab. \ref{tab:all-avg-errors} presents pose estimation errors in meters across four simulations (Sim 0 to Sim 3) using LiDAR and various sensor permutations, comparing SLAM, GPS, registration (Reg.), valid registrations and fused methods. 

To assess the reliability of the fused map for the "valid registration", we use two metrics: the fitness score of overlapping point clouds and the inlier Root Mean Square Error (RMSE). A well-registered, or "valid" frame is one whose quality measures are locally consistent and physically plausible. For each frame $i$, we compute the fitness $f_i$ (inlier ratio, higher is better) and RMSE $r_i$ (inlier distance error, lower is better). To account for local variations, we define envelopes over a sliding window of width $w$, namely the local maximum fitness and minimum RMSE:
$
    f^{env}_i = \max_{|j-i|<w/2} f_j, \hspace{3mm}    r^{env}_i = \min_{|j-i|<w/2} r_j.$

A frame $i$ is accepted if its metrics are close to these local optima:
$
    f_i \geq f^{env}_i - \delta_f, \hspace{3mm}  r_i \leq r^{env}_i + \delta_r,$
where $\delta_f$ and $\delta_r$ denote tolerance margins. Only frames satisfying all these conditions are regarded as well-registered.
 
GPS consistently shows the largest errors (7.988 -- 16.967 m), whereas SLAM maintains low errors (0.093 -- 0.164 m). Registration errors vary more widely (0.024 -- 59.212 m), but improve with additional sensors, such as in 4\_infra\_2\_agent configurations. The fused approach achieves the best performance, reducing errors to the millimeter range (0.009 -- 0.028 m in valid cases), demonstrating the benefits of combining SLAM and registration for robust localization in multi-agent setups.
\begin{table}[ht!]
    \centering
    \rowcolors{2}{gray!10}{white}
    \begin{tabular}{l|l|c|c|c|c|c|c}
        \rowcolor{gray!30}
        \textbf{Simulation} & \textbf{Permutation} & \textbf{SLAM} & \textbf{GPS} & \textbf{Reg.} & \textbf{Reg. (Valid)} & \textbf{Fused} & \textbf{Fused (Valid)} \\
        \hline
        Sim 0 & 4\_infra\_2\_agent & 0.164 & 7.988 & 3.929 & 1.046 & 1.634 & 0.016 \\
        Sim 0 & 4\_infra           & 0.164 & 7.988 & 3.615 & 0.875 & 0.019 & 0.016 \\
        Sim 0 & 2\_infra           & 0.164 & 7.988 & 2.525 & 0.565 & 0.019 & 0.016 \\
        Sim 1 & 4\_infra\_2\_agent & 0.145 & 11.203 & 2.715 & 2.330 & 0.021 & 0.020 \\
        Sim 1 & 4\_infra           & 0.145 & 11.203 & 1.488 & 0.996 & 0.028 & 0.028 \\
        Sim 1 & 2\_infra           & 0.145 & 11.203 & 2.101 & 1.903 & 0.019 & 0.019 \\
        Sim 2 & 4\_infra\_2\_agent & 0.093 & 16.967 & 5.252 & 4.225 & 0.019 & 0.014 \\
        Sim 2 & 4\_infra           & 0.093 & 16.967 & 32.530 & 32.697 & 0.021 & 0.020 \\
        Sim 2 & 2\_infra           & 0.093 & 16.967 & 5.796 & 3.200 & 0.016 & 0.012 \\
        Sim 3 & 4\_infra\_2\_agent & 0.102 & 14.541 & 3.275 & 1.864 & 0.016 & 0.015 \\
        Sim 3 & 4\_infra           & 0.102 & 14.541 & 8.658 & 0.043 & 0.019 & 0.012 \\
        Sim 3 & 2\_infra           & 0.102 & 14.541 & 6.989 & 0.026 & 0.023 & 0.009 \\
    \end{tabular}
    \caption{Pose estimation errors (in meters) across simulations and permutations. \textbf{Simulation}: ID; \textbf{Permutation}: Sensor config (e.g., \text{4\_infra\_2\_agent}: 4 infrastructure, 2 agent sensors); \textbf{SLAM}: SLAM error; \textbf{GPS}: GPS error; \textbf{Reg.}: Registration error; \textbf{Fused}: Error post SLAM-registration fusion. “Valid” columns: only well-registered frames. Errors: avg. distance between aligned estimated and ground-truth trajectories, measuring drift.}
    \label{tab:all-avg-errors}
\end{table}

\section{Conclusion} \label{ch:colclusion}
In this work, we use a cooperative multi-modal setup in autonomous vehicles, fusing point clouds from vehicle and infrastructure sensors like LiDAR and stereo cameras. Integrating GPS localization with ICP-RANSAC-enhanced SLAM, our approach corrects noisy GPS trajectories in urban settings, as shown in CARLA simulations across varied scenarios. Results indicate fused localization reaches millimeter accuracy (e.g., 0.009 -- 0.028 m in valid cases), far surpassing standalone GPS (7.988 -- 16.967 m errors) and SLAM (0.093 -- 0.164 m), underscoring multi-agent synergy for robust perception and mapping. Challenges like data delays, alignment issues, and scalability in traffic remain, requiring advanced calibration. Future efforts will investigate deep learning fusion, edge computing, and real-world validation to boost reliability in GPS-noisy environments.
\begin{credits}
\subsubsection{\ackname}
We gratefully acknowledge the contributions of Francesca Frederick, whose assistance with the Bachelor's thesis and graphics was invaluable.
\end{credits}
%
%
%
\bibliographystyle{splncs04}

\bibliography{mybibliography}
\end{document}